\documentclass[10pt,twocolumn,letterpaper]{article}

\usepackage{iccv}

\usepackage{times}
\usepackage{soul}
\usepackage{url}
\usepackage[utf8]{inputenc} 
\usepackage[small]{caption}
\usepackage{graphicx}
\usepackage{amsmath}
\usepackage{amsthm}
\usepackage{booktabs}
\usepackage{algorithm}
\usepackage{algorithmicx}
\usepackage{algpseudocode}
\usepackage{color}
\usepackage{enumitem}
\usepackage{multirow}
\usepackage{multicol}
\usepackage{enumitem}
\usepackage{svg}
\usepackage{color, colortbl}
\usepackage{pifont}

\usepackage[pagebackref=true,breaklinks=true,letterpaper=true,colorlinks,bookmarks=false]{hyperref}

\iccvfinalcopy 

\def\iccvPaperID{20} 

\ificcvfinal\fi

\begin{document}

\title{Multi-Task Hypergraphs for Semi-supervised Learning \\ using Earth Observations}

\author{
Mihai Pirvu\textsuperscript{1,2} \hspace{0.25cm}
Alina Marcu\textsuperscript{1,2} \hspace{0.25cm}
Alexandra Dobrescu\textsuperscript{1} \hspace{0.25cm}
Nabil Belbachir\textsuperscript{3} \hspace{0.25cm}
Marius Leordeanu\textsuperscript{1,2,3}
\thanks{Primary contact: Marius Leordeanu at leordeanu@gmail.com}
\vspace{1.5mm}\\
\textsuperscript{\rm 1}University Politehnica of Bucharest \\
\textsuperscript{\rm 2}Institute of Mathematics "Simion Stoilow" of the Romanian Academy\\
\textsuperscript{\rm 3}NORCE Norwegian Research Center
}

\maketitle
\ificcvfinal\thispagestyle{empty}\fi

\begin{abstract}
There are many ways of interpreting the world and they are highly interdependent. We exploit such complex dependencies and introduce a powerful multi-task hypergraph, in which every node is a task and different paths through the hypergraph reaching a given task become unsupervised teachers, by forming ensembles that 
learn to generate reliable pseudolabels for that task.
Each hyperedge is part of an ensemble teacher for a given task and it is also a student of the 
self-supervised hypergraph system. We apply our model to one of the most important problems of our times, that of Earth Observation, which is highly multi-task and it often suffers from missing ground-truth data. By performing extensive experiments on the NASA NEO Dataset, spanning a period of 22 years, we demonstrate the value of our multi-task semi-supervised approach, by consistent improvements over
strong baselines and recent work. We also show that the hypergraph can adapt unsupervised to gradual data distribution shifts and reliably recover, through its multi-task self-supervision process, the missing data for several observational layers for up to seven years. 
\end{abstract}

\vspace{-6mm}
\section{Introduction}

\begin{figure*}[t]
\centering
\includegraphics[width=0.9\linewidth]{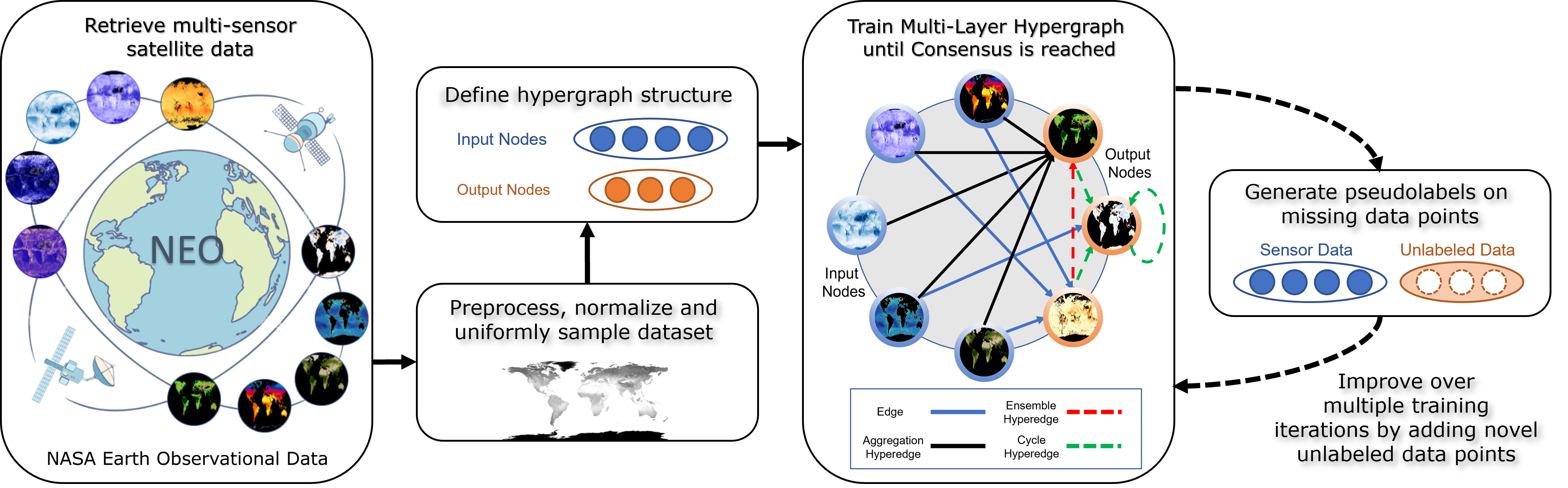}

\caption{\label{fig:main_figure} Overview of our Multi-Task Hypergraph for Semi-supervised Learning for Earth Observation, using available multi-sensor satellite data from the NEO platform (Planetary 2D maps logged since 2000). We sample data monthly, process it, and use multiple climate indicators to define the structure of our hypergraph: determine the input nodes (Earth observation layers that are always available), output nodes (layers to be predicted), and edge/hyperedge functions (modeling different input to output predictions). Multiple hyperedges will form ensembles to generate pseudolabels for data points with missing layers. We demonstrate that considering pseudolabels during training significantly improves learning over several iterations.} 
\vspace{-2mm}
\end{figure*}

The accuracy of deep learning is directly correlated with the quality and quantity of labeled data, especially in the context of supervised learning. Consequently, researchers focused their efforts on unsupervised pretraining methods or learning with limited supervision, by exploiting consistency among multiple tasks. However, a holistic semi-supervised solution that combines information from all available sensors and interpretation layers is still missing. The problem of learning for Earth Observation from satellite data, which we address in this paper, is increasingly important and it would greatly benefit from such a solution, as it includes data from a large number of observation layers, which are often missing due to faulty sensors.

Despite such real-world needs in applications that are by nature multi-task, there are still a very limited number of works that approach the problem from a holistic semi-supervised perspective. Most unsupervised learning papers focus on exploiting relations between only a limited number of tasks, such as relative pose, depth and even semantic segmentation~\cite{chen2019self, zhou2017unsupervised, ranjan2019competitive, bian2019unsupervised, gordon2019depth, yang2018unsupervised, tosi2020distilled, guizilini2020semantically, chen2019towards, stekovic2020casting} or modalities ~\cite{Hu_2019_CVPR, Li_2019_CVPR, zhang2017split, pan2004automatic, he2017unsupervised, zhao2018sound}. On the other hand, other methods that explore the relationships between many tasks do not fully address the semi-supervised learning scenario~\cite{zamir2018taskonomy, zamir2020robust, fifty2021efficiently}. 

\textbf{Main contributions: } We introduce several contributions, which come to address the limitations of the current semi-supervised multi-task learning literature by offering an integrated and robust solution to the problem:
1) We consider higher-order relationships between several tasks, which represent hyperedges in the hypergraph and can also act as
standalone models that 
predict any output layer (task) from potentially many input ones. 2) Second,
we propose to learn robust hyperedge ensembles, which combine multiple pathways reaching a given task node and become unsupervised teachers for the edges and hyperedges trained at the next semi-supervised iteration. 3) Third, we choose a problem that is by nature multi-task and also relevant for our society, that of Earth Observation, which is also novel in the multi-task semi-supervised learning literature. 

Our hypergraph approach to the aforementioned problem is related to well-known recent papers on semi-supervised learning using graphs ~\cite{ling2022semi,stretcu2019graph,goyal2020graph,feng2019hypergraph,
tudisco2021nonlinear, leordeanu2021semi, haller2021self, zamir2018taskonomy, zamir2020robust}, pretraining and task-specific fine-tuning ~\cite{girshick2014rich, huh2016makes, devlin2018bert, dosovitskiy2020image}, teacher-student distillation ~\cite{liu2019structured,wang2021knowledge,gou2021knowledge,wu2021real,
zhang2022topformer,gao2022fbsnet} and semi-supervised learning over multiple iterations
~\cite{marcu2020semantics,zhen2020joint, Yeo_2021_ICCV}. However, previous works are more limited with respect to the number of tasks considered, the lack of higher-order relationships between them or the inability to learn unsupervised. For more specific details of our approach, please go to Sec. \ref{sec:approach} onward. For a general overview see Fig.~\ref{fig:main_figure}.

\textbf{Climate impact.} One of the main contributions of our work from a real-world perspective is the chosen problem of Earth Observation, which is essential for better understanding Earth Climate. Climate forecasting for Earth Observation~\cite{neo} requires solutions from scientific, technical and social domains. Starting from the data, we are limited by the existing infrastructure: actual sensors and satellites, orbiting the Earth and making aggregated observations. One factor that influences the atmosphere and has an insufficiently understood impact is the \textit{aerosols}, which often occur from natural elements of the Earth's ecosystem, like sea salt or wildfires. These minute particles can be found at all altitudes and are a complex element of the climate model with both positive (aerosols encourage cloud formation which heats and cools the Earth) and negative (aerosols can reflect solar energy into space) effects~\cite{esa_aero}.
Ocean salinity is another factor that links to the causal chain of climate change, reshaping the density structure of oceans, which leads to a most alarming effect: sea-level changes~\cite{durack2014long}.
Compared to the traditional technique of weather forecasting from ~\cite{berndt2017transforming} or forest cover and road length, which are general prediction tasks exposed in~\cite{rolf2021generalizable}, we aim at integrating multiple climate factors, into our Multi-Task Hypergraph and exploit the relationships and consensus formed among different pathways to better understand and form cross-task correlations between multiple Earth layers. We experimentally show that for having a robust prediction of a given climate variable, we need multiple sensors that interact with each other and coordinate on different prediction paths.

\textbf{Remote sensing and AI for climate change.} With the help of aerial images, we can monitor the transitional stages of vegetation over a large period of time and determine the potential impacts of land and forest biomass. The work from~\cite{kwiatkowska2002merger} fuses different ocean satellite images together using classical Computer Vision techniques to provide a high-quality ocean dataset for scientific purposes. The work from~\cite{xiao2004modeling} uses various modalities, such as $CO_2$ and vegetation to predict the gross primary production (GPP) of a particular area, which is a determiner of the quality of life. The authors from~\cite{segovia2021does} use GNNs and satellite imagery to predict air quality in relation to COVID-19 spreading during the 2020 pandemic. The work from~\cite{segal2020cloud} used a CNN-based approach to detect clouds from satellite images, which are then further used to process other types of modalities where cloud removal is necessary. The model proposed in ~\cite{upadhyaya2022multi} is a classification framework for the identification of believer/denier attitudes expressed on social media platforms regarding the climate change problem. This work formulates environmental deviations as a multi-task learning problem within the natural language domain by combining stance detection and sentiment analysis. 
Other works~\cite{li2020nasa} used a multi-modal neural network to detect benthic tropical marine systems, such as coral reefs.
In a recent article~\cite{early2022scene}, the authors formulate the Land Cover Classification task as a Multiple Instance Learning regression problem to reduce the computational cost. 

While there is a rapidly increasing body of literature on AI for understanding the Earth Climate, our model is the first, to our best knowledge, that offers a holistic multi-view semi-supervised approach.
We argue that such an integrated research direction is needed for capturing the Earth System in its full interconnected complexity.

\begin{figure}[t]
\centering
\includegraphics[width=1\linewidth]{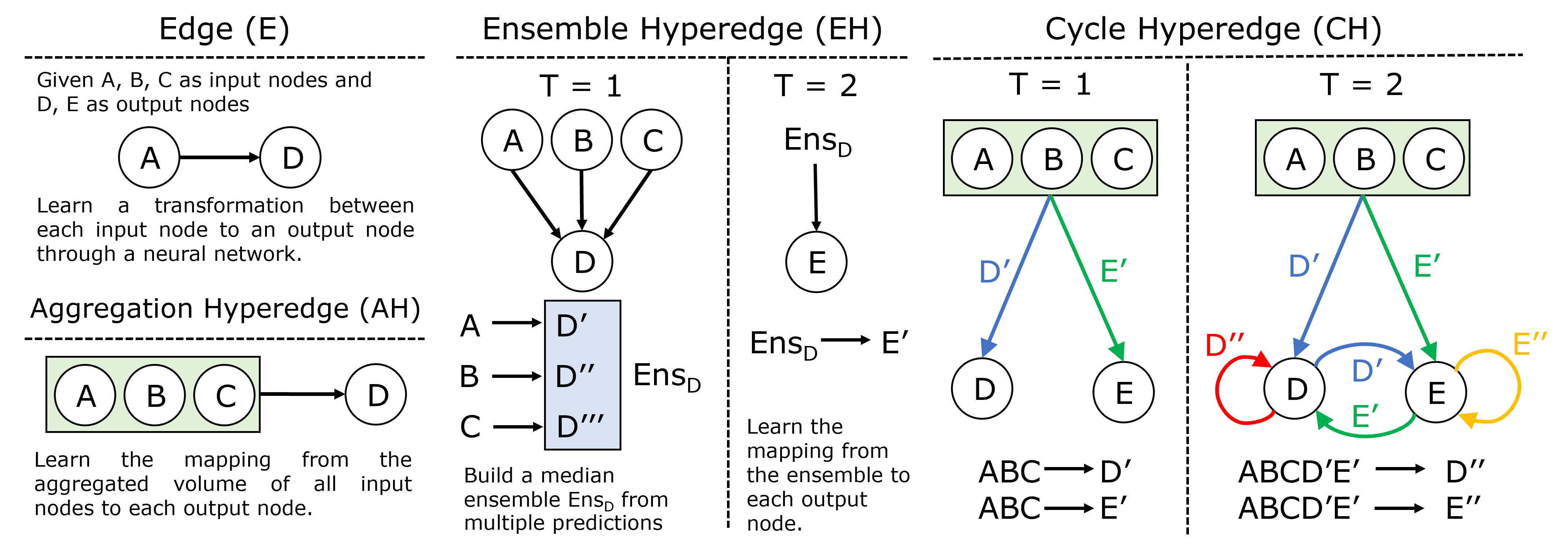}
\caption{\label{fig:hyperedges_types}  The basic processing units of the hypergraph are \textbf{direct neural links (DNL)}. Note that all arrows in the Figure are neural links. The neural links that connect an input node to an output node are simple edges (E), while the others, that connect multiple nodes to an output node form complex hyperedges. Given input nodes: A, B, C, and outputs D and E, we have four types of hyperedges of different complexities: pairwise edges (E), ensemble hyperedges (EH), aggregation hyperedges (AH) and cycle hyperedges (CH).}
\vspace{-2mm}
\end{figure}

\vspace{-2mm}
\section{Our Approach}
\label{sec:approach}

Our multi-task hypergraph (Fig. \ref{fig:main_figure}) has two sets of nodes, each node representing a different layer (view) of the world: known input nodes ($N_i$) and predicted output nodes ($N_o$). These nodes are connected by edges and hyperedges, such that, within each hyperedge (or edge) one or several nodes are transformed into a single output node. The basic processing units of the hypergraph are \textbf{direct neural links}, modeled by small U-Nets, which transform input nodes into an output one, as detailed in Sec. 4.
These links are used to construct the edges and different types of hyperedges in the hypergraph, as shown in Sec. 3. The hypergraph has many pathways that take information from input nodes to reach each a given output one, forming a large pool of layer candidates for each such output node.
We further combine these candidate views into ensembles, which automatically learn to produce a single robust final output for a given output node. We consider different types of ensemble models, going from simple to complex ones, as presented in Sec. 3. These ensembles act as unsupervised teachers for training the next generation edges and hyperedges, which, in turn form the next generation ensembles, within a continuous iterative self-supervised process.
The main steps of our Algorithm are described next.

\subsection{Algorithm}
\label{sec:algorithm}

\noindent \textbf{Step 1 - Initialize the hypergraph:} Using fully labeled data (Labeled Set $S_L$), we train the initial direct neural links $\mathbf{Y} \gets f_{l}(\mathbf{X})$, which take as input a volume of one or several layers $\mathbf{X}$, from one or several nodes, and transforms it into a single output layer $\mathbf{Y}$. The set of $f_{l}$ functions are our basic processing units, with which we construct all the edges and hyperedges (see Sec. \ref{sec:hypergraph_structure}). \\

\noindent \textbf{Step 2 - Learning hypergraph ensembles:} For each task (output node) $\mathbf{Y}^i$ in the hypergraph, multiple candidate outputs are produced by different pathways resulting from the many edges and hyperedges formed. Using strictly the Labeled Set $S_L$, and using the true label $\mathbf{Y}^{i}_{gt}$ as target output and all 
candidate pathway outputs as set of inputs, 
we learn ensemble functions $f^{i}_{ens}([\mathbf{Y}^{i}_1, \mathbf{Y}^{i}_2, \dots, \mathbf{Y}^{i}_C],
\mathbf{w}_{ens})$, to produce a single output $\mathbf{Y}^{i}_{ens}$, from $C$ candidates, by learning to minimize 
the L2 loss:  $\|\mathbf{Y}^i_{ens}- \mathbf{Y}^i_{gt}\|_2$. \\ 

\noindent \textbf{Step 3 - Generate pseudolabels:} At iteration $k$, on the Unlabeled Set $S_U$, we pass information (inference) from input nodes $\mathbf{X_i}$ all the way to the output ones, using the set of learned edges, hyperedges and their ensembles, to produce final pseudolabels $\mathbf{Y}^{(i,k)}_{ens}$ for each output node $i$. \\

\noindent \textbf{Step 4 - Semi-supervised learning:} We add the Unlabeled Set $S_U$ with the newly generated pseudolabels to the fully Labeled Set $S_L$, and retrain from scratch all the neural links used to form edges and hyperedges, by minimizing the total sum of squared error loss over all links, which map inputs to outputs directly and can be efficiently trained independently: 

\vspace{-5mm}
\begin{equation}
L= \sum^{N_o}_{i=1}(\sum_{\mathbf{X}\in S_L}\|f_{l}(\mathbf{X})-\mathbf{Y}^{i}_{gt}\|^2_2+\sum_{\mathbf{X}\in S_U} \|f_{l}(\mathbf{X})-\mathbf{Y}^{(i,k)}_{ens}\|^2_2)
\end{equation}

After retraining the links (which implicitly updates all edges and hyperedges), we return to Step 2 for a new semi-supervised learning iteration $k \gets k+1$, until convergence.

\vspace{-2mm}
\section{Multi-Task Hypergraph Structure}
\label{sec:hypergraph_structure}

Below we present the different types of hyperedges and ensemble models that define our hypergraph:

\subsection{Hyperedges}
Within the structure of our hypergraph, we form four types of hyperedges: 
\textbf{1. Edges (E):} They learn a transformation between each input node to an output node through a single neural network link, for a total of $N_i \times N_o$ edges. \textbf{2. Ensemble Hyperedges (EH):} All the edge (E) predictions for an output node are combined using the pixelwise median ensemble function, to obtain an output map for each of the $N_o$ output nodes (T=1). Then from each such output map at T=1, we learn a separate link for each of the remaining $N_o-1$ output nodes (T=2). Thus, each output receives (at T=2) $N_o-1$ candidates from such EH edges, which are $N_o(N_o-1)$ in total number.
\textbf{3. Aggregation Hyperedges (AH):} They concatenate all input nodes and use a single direct neural link (with a modified input volume to include all input layers) to directly produce an output node. There are in total $N_o$ AH hyperedges in the hypergraph.
\textbf{4. Cycle Hyperedges (CH):} They concatenate all the $N_i$ input nodes together with the $N_o$ outputs of all AH hyperedges (T=1) to produce the output node at T=2 by another direct link (with correctly modified input volume). There is one such CH hyperedge for each of the $N_o$ nodes.

\subsection{Multi-path Ensemble Models} 
\label{sec:ensembles}

Different from previous work, which used simpler non-parametric ensemble models for generating pseudolabels at output nodes, we introduce ensembles that learn how to combine the different candidates at each output node layer and get retrained with each semi-supervised iteration.
We propose several such models, starting from simpler ones that linearly combined the candidates using fixed weights, to more powerful ones, using neural nets, that learn to produce directly a final output map from several candidates. 

\begin{figure}
  \centering
  \includegraphics[width=\linewidth]{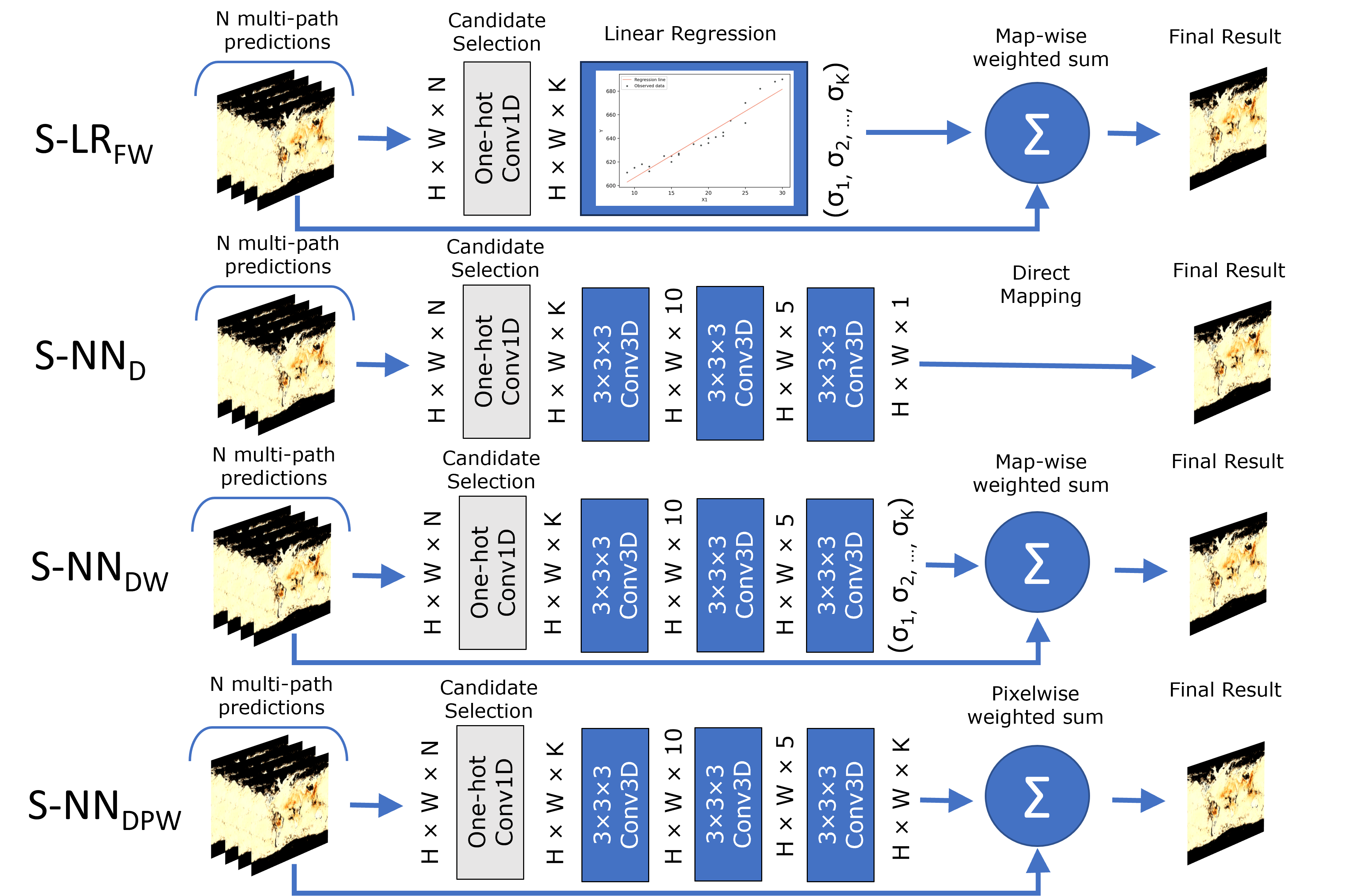}
  \caption{Ensemble Architectures.
  We introduce four types of ensembles, all with an initial learnable candidate selection model, which keeps only the relevant candidates before combining them:  S-LR$_{FW}$ learns one fixed weight per candidate. S-NN$_{DW}$ dynamically outputs a weight per candidate, depending on the input, while S-NN$_{DPW}$ dynamically outputs a weight for each pixel of each candidate. Instead of linearly combining the candidates S-NN$_{D}$ learns a direct non-linear mapping from candidates to output. All are learned end-to-end.  
  }
\label{fig:conv3ds-neural-aggregator}
\vspace{-4mm}
\end{figure}

We also introduce an automatic input candidate selection procedure for all ensemble models, which learns a non-linear weight function $\frac{1}{1+\exp{(-\alpha_c)}}$ per input candidate channel, by which we multiply each candidate before entering the ensemble model. Being between 0 and 1, this weight effectively turns on or off a given candidate, and thus learns to keep only the relevant ones, for improved robustness.
In Fig. \ref{fig:conv3ds-neural-aggregator} we show the different learned selection and ensemble models, which are of the following types: \textbf{S-L$_{FW}$} is a linear model with selection, which learns to linearly combine the channel candidates by a fixed set of weights $w_c$, such that the final weight per candidate is effectively 
$w_c\frac{1}{1+\exp{(-\alpha_c)}}$. \textbf{S-NN$_{DW}$:} a neural net model with selection, which learns to produce a dynamic weight $w_c$ per channel (that changes dynamical depending on the input), such that the final output is obtained by linearly combining the channels multiplied by the dynamic weights. \textbf{S-NN$_{DPW}$:} a neural net model with selection, which learns to produce a dynamic pixelwise weight $w_{ci}$ per channel $c$ and pixel $i$, such that the final output is obtained by linearly combining the pixels of each input channels multipled with the corresponding output pixelwise weights. \textbf{S-NN$_{D}$:} a neural net with selection, which learns a direct transformation between all input candidate layers to the final output. Note that, in principle, S-NN$_{D}$ is the most flexible, non-linear model. However, since the ensembles are trained on the limited Labeled Set $S_L$, where true target values are available, a simpler weighted combination of the input channels (fixed, dynamic or pixelwise dynamic) could be more robust. What changes in the training of these ensembles during semi-supervised iterations are their input candidates  $\mathbf{Y}_c$ - produced every iteration by updated edges and hyperedges. Also note that all these models, including the initial selection module, are trained end-to-end.

\vspace{-2mm}
\section{Implementation Details}
\label{sec:implementation}

We used PyTorch~\cite{NEURIPS2019_9015} as our main Python Machine Learning framework, alongside~\cite{ngclib} for the semi-supervised iterative graph-ensemble library. Below we detail the architecture of the direct neural links (DNL): \\

\vspace{-3mm}
\noindent \textbf{Direct neural link architecture (DNL):}
This link is the building block used to create all edges and hyperedges. It is a small U-Net of 1M parameters, the same as the one used in NGC~\cite{leordeanu2021semi}. It has an encoder branch with 3 down-sampling blocks, a central part with summed dilated convolutions and increasing dilation rates of 1, 2, 4, 8, 16 and 32, and a decoder with three up-scaling blocks. All kernels are of size $3 \times 3$, and each dilated convolution produces a set of 128 activation maps. Each downsampling block has 2 convolution layers with stride 1, followed by a convolution layer with (2, 2) strides to halve the feature maps. Similarly, each up-scaling block consists of a transposed convolution layer, a feature map concatenation with the corresponding map from the downsampling layers and 2 convolutional layers with stride 1. The number of feature maps is 16, 32 and 64 for the downsampling blocks and the opposite order for the upsampling blocks. Across the entire network, we only use ReLU as the activation function. \\

\vspace{-3mm}
\noindent\textbf{Training details.} We used the same training setup for all neural links that form the edges and hyperedges. All learning tasks are treated as regression problems, with standard L2-norm loss function. The networks are optimized for 100 epochs, with the Adam optimizer, an initial learning rate of 0.01 and a learning rate scheduler, that after an initial 10 epochs, divides by 2 the current learning rate if there is no improvement for 5 consecutive epochs. \\

\vspace{-3mm}
\noindent\textbf{Training infrastructure.} WE used a medium-sized server with 8x NVIDIA 2080ti, a Xenon Gold 5218 CPU and 512GB of RAM. On average, training one edge (1 input and 1 output) took between 1-2h, depending on the iteration.

\vspace{-2mm}
\section{Experimental Analysis}

\textbf{NASA Earth Observations.} We use the multi-layer representations from the Earth Observations NEO Dataset provided by NASA~\cite{neo}. Satellites continually orbit the globe, collecting information about Earth in five main domains: Atmosphere, Energy, Land, Life and Ocean. In order to maximize the amount of overlap between multiple layers in time we sample monthly data points for each representation spanning the period from 2000-2022. The NEO dataset consists of multiple satellite observations, each of them being a monthly aggregate of that particular representation. Although we tried to maximize the amount of overlap between the representational layers, we faced challenges due to missing data, as some sensors began or ended at different points in time. 

\textbf{Graph configuration for NEO data.} We use the following 12 representations as input nodes: Normalized Difference Vegetation Index ($NDVI$), Snow Cover ($SnowC$), Land Surface Temperature Day and Night ($LSTD$, $LSTN$), Cloud Optical Thickness ($CLD_{OT}$), Cloud Particle Radius ($CLD_{RD}$), Cloud Fraction ($CLD_{FR}$), Cloud Water Content ($CLD_{WP}$), Nitrogen Dioxide ($NO_{2}$), O-zone layer ($OZONE$), Chlorophyll ($CHLORA$), Sea Surface Temperature ($SST$). Our multi-task learning graph predicts the following 7 types of output representations: Aerosol Optical Depth ($AOD$), Active Fires ($FIRE$), Water Vapor ($WV$), Anomaly Layers for LSTN and LSTD ($LSTN_{AN}$, $LSTD_{AN}$), Leaf Area Index ($LAI$) and Carbon Monoxide ($CM$). Visual samples are provided in Figure~\ref{fig:main_figure}. All maps corresponding to raw sensor values, from the NEO platform~\cite{neo}, were rescaled to a resolution of $1080\times540$ and normalized in a range of $[0:1]$. As explained in Sec. \ref{sec:hypergraph_structure}), we have $N_i \times N_o = 84$ single edges (E), $N_o \times (N_o-1)=42$ ensemble hyperedges (EH), $N_o = 7$ aggregation hyperedges (AH) and $N_o = 7$ cycle hyperedges (CH). Please see again Figure \ref{fig:hyperedges_types} for a visual presentation of these types of hyperedges.

\textbf{Dataset Split.} The Labeled Set, used for supervised learning (Iteration 1) of all neural links, consists of the first 119 months. The following 30 months, the \textbf{Test Set}, are used for evaluation. The next remaining 62 months (until 2022) represent the Unlabeled Set, used for semi-supervised learning (Iteration 2). The main steps of the learning algorithm are presented in Sec. \ref{sec:approach}.  

\begin{figure*}[h!]
  \centering
  \includegraphics[width=1\linewidth]{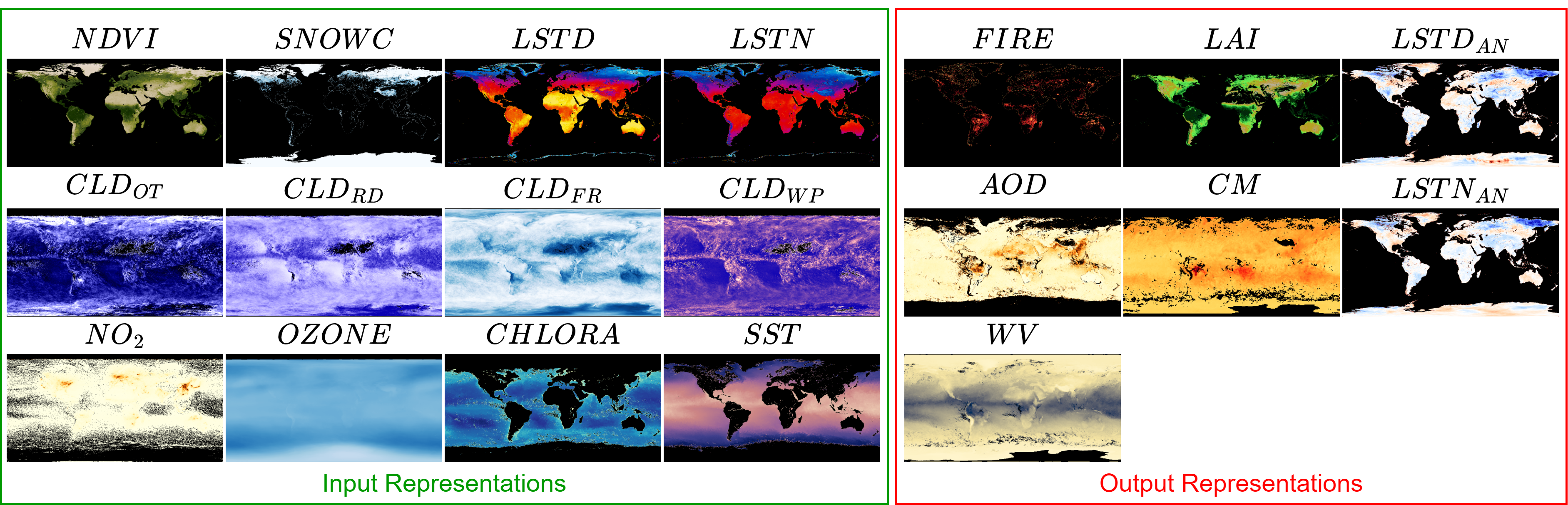}
  \caption{Visual samples from the NEO dataset for each of our input and output representations. Each layer comes from 4 main categories: Atmosphere, Land, Life, or Ocean.}
\label{fig:vis_samples_dataset}
\end{figure*}

\begin{figure}[h!]
  \centering
  \includegraphics[width=1\linewidth]{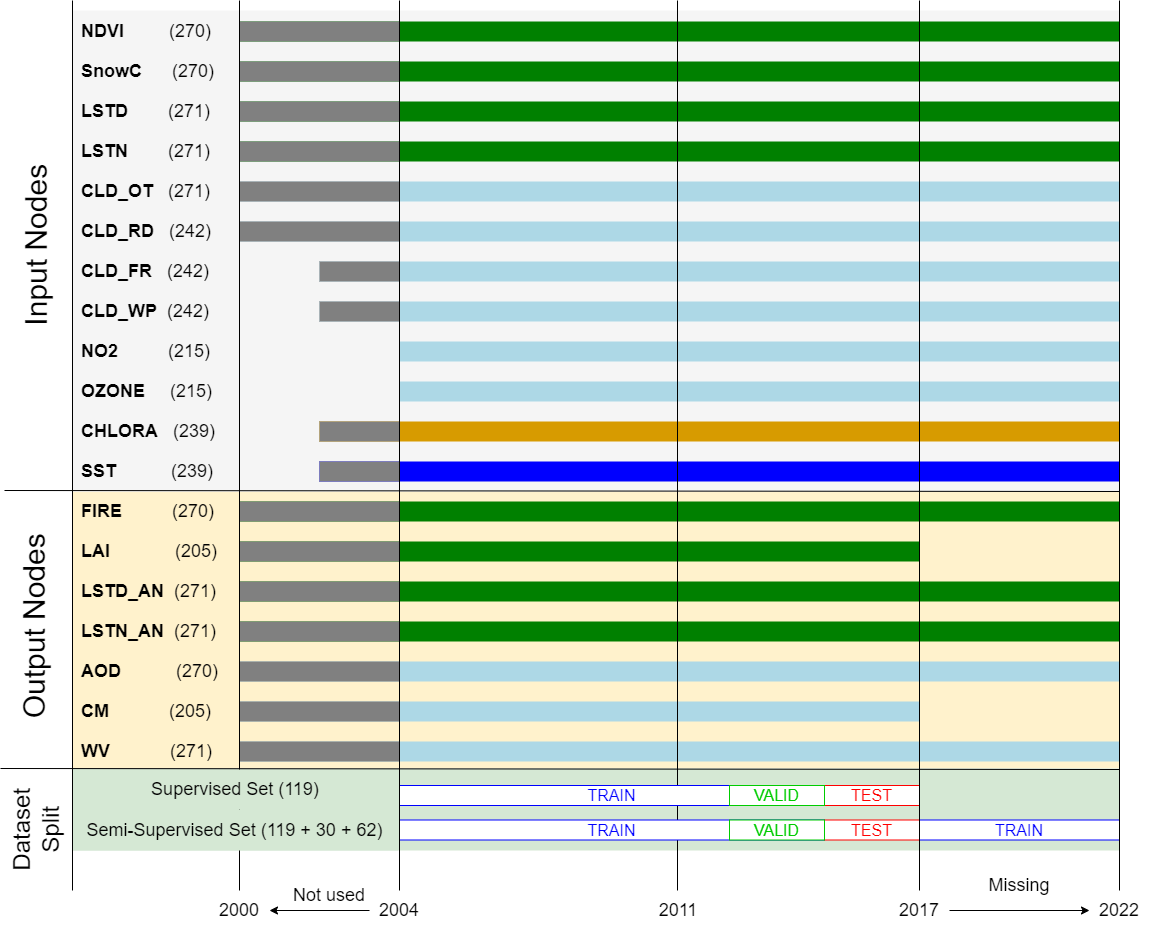}
  \caption{Graphical summary of the aggregated monthly observations available in NEO Database for our 12 input nodes and 7 output nodes. Best viewed in color. Climate sensors can be divided into 4 main categories: Land (dark green), Atmosphere (Light blue), Ocean (Dark blue) and Life (Light orange). Hard-colored gray bars indicate available data points.}
\label{fig:nasa_dataset_timespan}
\end{figure}

\textbf{Dataset timespan details.}
Visual samples from the dataset show that some layers can have large areas of missing information (blank), depending on what a particular sensor is focusing on: e.g. land (sparse) or atmosphere (dense), which makes learning transformations between each of these layers cumbersome. The visual differences between each of the selected layers from the dataset are highlighted in Figure~\ref{fig:vis_samples_dataset}.
Our selection of Earth layers from NEO was not arbitrary. The data spans the period between 2000-2022, but as Figure~\ref{fig:nasa_dataset_timespan} shows, there are still many missing data points. Although we tried to maximize the amount of temporal overlap between multiple representational layers, we still had some missing data points for observations such as $NO_{2}$ and $OZONE$ where the data collection started in 2004, and for $LAI$ and $CM$, where it stopped in 2017. This lack of data points motivates our choice of dataset split - since we can use our hypergraph to generate pseudolabels and then use them in the iterative semi-supervised learning procedure. In the lower part of Fig.~\ref{fig:nasa_dataset_timespan}, we present the dataset split. Train indicates $S_L$, the period used for supervised training (Iteration 1). Where data is missing (Unlabeled Set $S_U$), we generate pseudolabels and use them alongside ground truth for semi-supervised training (Iterations 2 and 3). 

\textbf{Evaluation metrics.} Our main metric is the \textit{Relative Performance Improvement} (RPI) with reference to the $L2$ score, against a fixed baseline: $RPI\% = (\frac{L2(y_{baseline}, gt)}{L2(y_{pred}, gt)} - 1) * 100$. We use this due to scaling issues that arise when using absolute values for multiple tasks, each of them having its own scale. The fixed baseline is the top-performing supervised edge (E) for each task. Furthermore, the $L2$ metric is computed as $L2 (y, gt) = \frac {\sum{((y - gt) \cdot mask) ^ 2}} {\sum{mask}}$, where $mask = (gt \ne NaN)$. Since satellite maps contain invalid values in the regions where the sensors could not pick any reliable information, we only train and evaluate on the valid ones. Here, $mask$ refers to a binary map, with a constant of $1$ only for valid data. For clarity and analysis of the overall trends, we also report the 
\textit{Average Relative Performance Improvement} (ARPI), that is the average of the RPI across all tasks combined.

\subsection{Comparison to multi-task baselines}

\begin{table}
    \setlength{\tabcolsep}{3.5pt} 
    \centering
    \begin{tabular}{lrr}
        \toprule
        Method & Iteration 1 (sup) & Iteration 2 (semi-sup) \\
        \toprule
        MTE-1M & -37.91 & -33.83 \\
        MTE-70M & \underline{-15.22} & \underline{-10.15} \\
        MTE-140M & -37.33 & -39.89 \\
        HG-140M & \textbf{7.77} & \textbf{8.85} \\
        \bottomrule
    \end{tabular} 
    \caption{Quantitative comparison in ARPI ($\uparrow$) to multi-task baselines (MTE) with different numbers of parameters, varying from 1M (similar to one edge) to 140M (similar to the whole hypergraph), for both supervised and semi-supervised scenarios.
    MTE concatenates all input nodes and transforms them into all output ones. Our hypergraph (HG) is vastly superior, while the highly inter-connected MTE clearly overfits. This justifies our multi-path hyperedge ensemble approach.
    }
    \label{tab:baselines-comparison}
\end{table}

One of the main objectives of multi-task learning is to leverage multiple tasks simultaneously in order to improve generalization performance across all tasks. Due to the problem of negative transfer ~\cite{jiang2023forkmerge}, it is still unclear how to combine these tasks without hurting the overall performance. Our experiments here also prove this point (Tab. \ref{tab:baselines-comparison}). 
We consider 3 multi-task baselines to compare against, referred to as MTE, with different number of parameters: 1.1M (the same as one edge), 70M and 140M (the same as the whole hypergraph).
The MTE is a single U-Net model, having a similar basic structure as our neural links (but different number of filters), which takes all 12 input layers and transforms them into all 7 output ones. The results show that such a model, which maps input to output, and lacks the robust multi-path hypergraph structure is not able to generalize when having very limited labeled data. By varying the number of parameters from very small to large, we show that this is not a problem of model capacity, but rather of model structure, which is exactly one of the main issues addressed in this paper.
In the semi-supervised case, the hypergraph and the MTE models are all trained on the same available ground truth together with the
pseudolabels they generate on the Unlabeled Set. 

\subsection{Impact of the unsupervised ensemble teachers}

Before proceeding with the iterative semi-supervised learning scenario, we first evaluate the performance of the ensemble models
proposed, to better understand the importance of learning to combine the candidates for each given output node. We also compare our ensembles with relevant published methods, such as the powerful Extreme Gradient Boosting method~\cite{xgboost}, in two variants, linear XGB-L and tree-based XGB-T, and recent work on multi-task graphs, based on non-parametric ensembles, such as NGC~\cite{leordeanu2021semi} (simple mean ensemble without selection) and CShift~\cite{haller2021unsupervised} (pixelwise kernel-based ensemble). 
We also add to the pool a simple ensemble, S-Mean, which simply averages the candidate layers after our automatic candidate selection procedure. In these tests, We let all approaches (including ours) have access to the same candidate layers, but using only Edges and Ensembles Hyperedges (EH). For a study of how the addition of the more complex hyperedges (AH and CH) boosts performance, see Sec. \ref{sec:hyperedges_vs_edges}.
The results, reported in Tab.~\ref{tab:ensembles_table}, show that the selection procedure improves performance (e.g. see S-Mean vs. NGC and
CShift), while an increased degree of model flexibility also improves performance. However, the top performer is not the most flexible full neural net mapping (S-NN$_D$) from input layers to output, but the next most flexible
model that produces dynamic pixelwise linear weighting
of the input layers (S-NN$_{DPW}$). In general simpler linear models (S-L and XGB-L) prove robust in these experiments, due to the limited unlabeled training data on which richer models can esily overfit.

\vspace{-2mm}
\begin{table}
    \setlength{\tabcolsep}{1.5pt} 
    \centering
    \begin{tabular}{lccccccccc}
        \toprule
        Ens. Type & (1) & (2) & (3) & (4) & (5) & (6) & (7) & ARPI ($\uparrow$)\\
        \midrule
        NGC \cite{leordeanu2021semi} & 0.44 & 12.97 & 12.15 & 5.44 & -50.2 & -39.0 & 2.21 & -8.01 \\
        CShift \cite{haller2021unsupervised} & 1.86 & 13.07 & 8.66 & 6.64 & -43.5 & -28.0 & 2.65 & -5.52 \\        
        XGB-L~\cite{xgboost} & 4.27 & 13.49 & 12.14 & \underline{9.61} & -2.71 & 1.24 & 5.95 & 6.28 \\
        XGB-T~\cite{xgboost} & \underline{6.13} & 6.76 & 5.75 & 7.24 & \textbf{2.29} & 4.83 & \underline{6.47} & 5.64 \\
        \midrule        
        S-Mean & 3.31 & \underline{14.71} & \textbf{12.42} & 8.97 & -9.50 & -2.48 & 5.93 & 4.77
        \\
S        S-LR$_{FW}$ & 4.85 & 11.15 & 8.62 & 8.45 & 0.21 & \underline{4.90} & 5.84 & 6.29
        \\
        S-NN$_{D}$ & \textbf{6.67} & 13.20 & 10.56 & \textbf{9.68} & 0.69 & 3.57 & \textbf{7.86} & \underline{7.46}
        \\
        S-NN$_{DW}$ & 4.79 & \textbf{14.80} & \underline{12.39} & 9.50 & 0.25 & \textbf{4.91} & 6.03 & \textbf{7.52}
        \\
        S-NN$_{DPW}$ & 5.74 & 6.51 & 11.21 & 8.26 & \underline{1.33} & 5.62 & 4.98 & 6.24
        \\
        \bottomrule
    \end{tabular}
    \caption{Evaluation on the Test Set of different ensemble models, for each output node separately (using RPI metric) and overall (ARPI metric). Ours have names starting with "S-" (denoting our proposed automatic candidate selection procedure). The best numbers are bolded, while the second best are underlined. We report relative performance (in percentage) for each output node in order (1) - $AOD$, (2) - $CM$, (3) - $FIRE$, (4) - $LAI$, (5) - $LSTD_{AN}$, (6) - $LSTN_{AN}$, (7) - $WV$, with reference to the best-performing edge for each task.}
    \label{tab:ensembles_table}
\end{table}

\subsection{Impact of semi-supervised learning iterations}
\label{subsec:learning-over-multiple-iterations}

\begin{figure}
  \centering
  \includegraphics[width=1\linewidth]{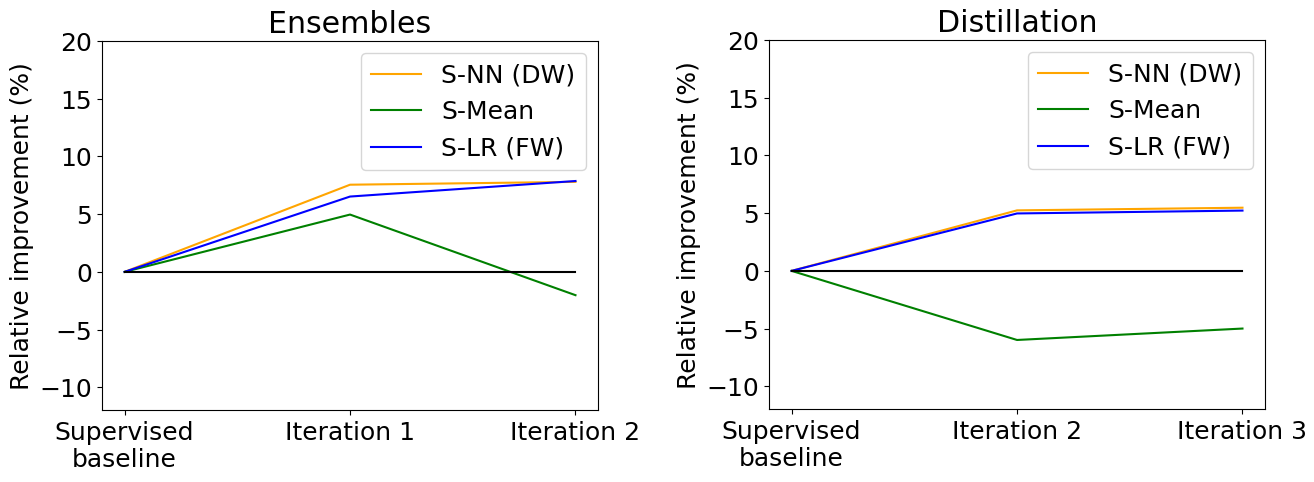}
  \caption{Iterative semi-supervised learning results -- Average Relative Performance Improvement over all test data points and output nodes (higher is better). Note that the simple mean (used by previous works) is not able to improve after semi-supervised learning. However, both learned ensemble functions improve significantly and keep improving slightly even at the second semi-supervised iteration.}
\label{fig:iter_results}
\end{figure}

For the scenario of multiple iterations of semi-supervised learning, when unlabeled data is added, we experiment with three ensemble models for generating pseudolabels, ranging from simple to more complex: S-Mean, S-LR$_{FW}$ and S-NN$_{DW}$. The first ensemble gives equal weights to the candidate layers, the second has different but fixed weights, while the third linearly combines the pixels of the input layers with weights that change dynamically, depending on the input. The same ensemble is used for each iteration of semi-supervised learning, resulting in three different hypergraph models (each with its own ensemble type) - so we can evaluate the impact of these types of ensembles completely independently, through several semi-supervised iterations. In Figure~\ref{fig:iter_results} 
we show the quantitative results (using the ARPI metric, over all tasks). The Distillation plot shows the average of the best-performing edges per task, after each semi-supervised retraining iteration (e.g., the distilled iteration 2 edges are the ones retrained on the ground truth labels + pseudolabels obtained at Iteration 1).
Interestingly, the results show that very simple models, such as S-Mean are not able to sustain the effectiveness of semi-supervised learning over several iterations, while models of medium complexity (such as S-LR) could be almost as efficient as more powerful ones based on neural nets (S-NN).
Note that the improvement in the distilled single best edge due to our semi-supervised hypergraph training means that there is in fact no added time or memory cost at test time. The hypergraph can be efficiently used only at training time.

\subsection{Improving over temporal distribution shifts}

\begin{figure}
  \centering
  \includegraphics[width=1\linewidth]{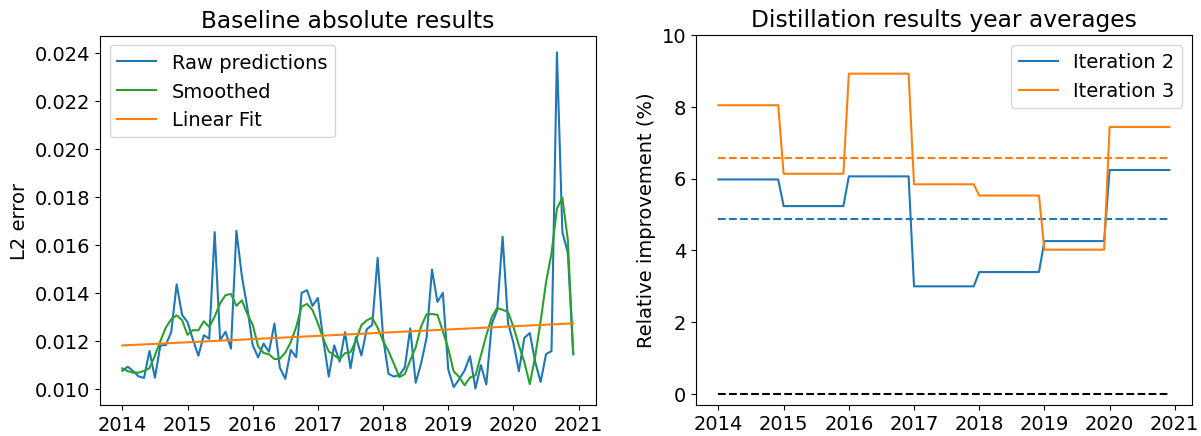}
  \caption{Prediction errors over seven years, averaged over 5 out of the 7 tasks. \textbf{Left-side}: L2 errors of the baseline supervised prediction. The linear fit shows a slight increase in error, which indicates that Earth Observation distribution data gradually shifts over time.
  \textbf{Right-side}: Relative error improvement over the supervised case, given by the first and second iterations of our semi-supervised learning. Note the significant improvements brought by these iterations (solid lines are averages per year and dotted lines are averages over seven years). 
}\label{fig:overtime_results}
\vspace{-4mm}
\end{figure}

We analyze the evolution of prediction errors across 7 years. These tests are conducted on 5 out of the 7 tasks, for which we had continuous ground truth labels for evaluation. Results are reported in Figure~\ref{fig:overtime_results}.
In the left plot, we start with the average L2 error of the edge baselines (trained fully supervised) across all tasks (blue line). We observe that the error follows a sinusoidal curve, which correlates precisely with the different times of the year (seasons).
We also add a smoothed and linear fit variant and observe that the error increases as we stray away from the Labeled Set used for training the edges. The relative error increase between 2014 and 2021 is 7.79\%. Also, there is an increase in the oscillation amplitude as well. These changes in error average and variance suggest that the data distribution is gradually shifting, probably due to changes in the Earth's climate. It also motivates the need to be able to continuously track such changes in an unsupervised way, when real ground truth data is missing.

The right-side plot from Figure~\ref{fig:overtime_results} presents the distillation results, on the same single edges as the first plot, following the iterative training done on S-NN$_{DW}$, but also averaged across each year to cancel the seasonal changes.
We present relative errors to the baseline, with 0 being the result of the first plot. The only difference between the first iteration and the
next one are the pseudolabels added. We observe that, on average and across all tasks, adding pseudolabels not only improves the predictions but also stabilizes the distribution shift across time. We also observe that Iteration 3 improves over Iteration 2, but the added value is lower.

\subsection{Improving temporal consistency}

We introduce a time-consistency metric, as the inverse of the prediction variance at a given geographic location (pixel) across a short time window centered around a given moment in time.
True values at a given location, do not fluctuate strongly, on average, over short periods of time - so the lower this variance, the more consistent and trustworthy the results should be, in general. For each particular output task, we compute all such variances for all locations and around all moments in time, in the test data set - and take the average as the overall consistency for a given task.

Our experiments reveal that the semi-supervised learning iterations (when pseudolabels are added) significantly improve, for the same edges, not only average performance (ARPI) but also the temporal consistency, for all tasks (Table~\ref{tab:consistency_table}). We compare the distilled version at Iteration 3 of the same network against its supervised counterpart (baseline). 

It should be noted that better consistency does not necessarily mean better performance, but rather a complementary score to performance. A constant prediction of 0 would have perfect consistency, however, the performance of the prediction would be bad. This metric can help us choose between two classifiers with similar performance but with different consistency scores. For a video representation of consistency maps, see \cite{yt_consistency_video}.

\begin{table}
    \setlength{\tabcolsep}{3pt} 
    \centering
    \begin{tabular}{llccc}
        \toprule
        Method & Ensemble & Iteration & Variance ($\downarrow$) & ARPI  ($\uparrow$)\\
        \toprule
        \multirow{4}{*}{Best Edge} & - & 1 & 0.051 & 0.0\\
        & S-Mean & 2 & \textbf{0.026} & -5.92\\
        & S-LR$_{FW}$ & 2 & 0.037 & \underline{5.02}\\
        & S-NN$_{DW}$ & 2 & \underline{0.036} & \textbf{5.21}\\
        \bottomrule
    \end{tabular}    
    \caption{We measure the consistency of the best-performing edge averaged over all output nodes and the ARPI before and after iterative semi-supervised learning using the S-Mean, S-LR$_{FW}$ and S-NN$_{DW}$ ensembles. Iteration 1 - supervised edge, Iteration 3 - final semi-supervised edge. We show significant improvement in the consistency metric for all the hypergraph distilled variants.}
    \label{tab:consistency_table}
\end{table}

\vspace{-2mm}
\subsection{Impact of complex hyperedges vs edges}
\label{sec:hyperedges_vs_edges}

\begin{table}
    \setlength{\tabcolsep}{2pt} 
    \centering
    \begin{tabular}{lccc}
    \toprule
	Type & ARPI ($\uparrow$) & ARPI w/ hyperedges ($\uparrow$) & Improvement \\
    \midrule
	S-Mean & 4.77 & \textbf{6.67} & 1.995 \\
	S-LR$_{FW}$ & 6.29 & \textbf{7.05} & 0.889 \\
	S-NN$_{DW}$ & 7.52 & \textbf{8.75} & 1.225 \\
    \bottomrule
    \end{tabular}
    \caption{Relative improvements for ensembles constructed without complex hyperedges (reported in Table~\ref{tab:ensembles_table}) vs. the ensembles considering the two complex hyperedges types added in the candidates set. The significant improvement demonstrates the potential power of these complex hyperedges.}
    \label{tab:hyperedges_table}
\end{table}

We evaluate the introduction of the more complex cycle and aggregation hyperedges 
(as presented in Fig.~\ref{fig:hyperedges_types} and Sec. ~\ref{sec:hypergraph_structure}), to study the effect of these more robust candidates in the selection and aggregation ensemble process. For the previous tests, we used a pool of 18 candidates for each output node (12 edges(E) and 6 ensembles hyperedges EH) when creating our ensembles. 
In  Tab.~\ref{tab:hyperedges_table} we show the results where we add these two hyperedges to the selection and aggregation pool, resulting in an ensemble of 20 edges, for each output node. Also, when we evaluate the performance of each edge or hyperedge independently for each task, one or both of these two complex hyperedges are the top performers on 5 out of 7 tasks. 
This clearly demonstrates the power of capturing higher-order relationships between tasks, but the power of the ensembles remains of great importance, as shown by our comparisons between different ensemble models and to single large nets MTE baselines,

\vspace{-2mm}
\section{Conclusions}

Our novel semi-supervised learning approach learns from robust ensemble teachers in a multi-task hypergraph of neural nets. Our experiments prove several points: 1) the addition of hyperedges that capture higher-order relationships between tasks can bring a strong boost to the performance of the ensemble teachers and the overall multi-task system. 2) Learning how to select and combine the relevant candidates also significantly boosts performance and improves the effectiveness of unsupervised learning. Also, when faced with very limited labeled data, simpler (e.g. selection + linear regression), but not simplistic (plain average) ensemble functions, can be almost as powerful as complex ones (neural nets). 3) Grouping the set of tasks into separate hyperedges and then combining them through ensembles can be much more robust than considering all tasks together in a single dense neural net. 4) Our multi-task hypergraph is also able to handle, self-supervised, multi-task distribution shifts that take place over longer periods, which makes it suitable for further development in the context of semi-supervised continual learning.

\noindent \textbf{Acknowledgements:} This work was funded in part by UEFISCDI, under Projects EEA-RO-2018-0496 and PN-III-P4-ID-PCE-2020-2819 and by a Google Research Gift.

{\small
\bibliographystyle{ieee_fullname}
\bibliography{egbib}

\begin{thebibliography}{10}\itemsep=-1pt

\bibitem{berndt2017transforming}
Emily Berndt, Andrew Molthan, William~W Vaughan, and Kevin Fuell.
\newblock Transforming satellite data into weather forecasts.
\newblock {\em Eos}, 98(3):26, 2017.

\bibitem{bian2019unsupervised}
Jiawang Bian, Zhichao Li, Naiyan Wang, Huangying Zhan, Chunhua Shen, Ming-Ming
  Cheng, and Ian Reid.
\newblock Unsupervised scale-consistent depth and ego-motion learning from
  monocular video.
\newblock In {\em Advances in Neural Information Processing Systems}, pages
  35--45, 2019.

\bibitem{chen2019towards}
Po-Yi Chen, Alexander~H Liu, Yen-Cheng Liu, and Yu-Chiang~Frank Wang.
\newblock Towards scene understanding: Unsupervised monocular depth estimation
  with semantic-aware representation.
\newblock In {\em Proceedings of the IEEE Conference on Computer Vision and
  Pattern Recognition}, pages 2624--2632, 2019.

\bibitem{xgboost}
Tianqi Chen and Carlos Guestrin.
\newblock {XGBoost}: A scalable tree boosting system.
\newblock In {\em Proceedings of the 22nd ACM SIGKDD International Conference
  on Knowledge Discovery and Data Mining}, KDD '16, pages 785--794, New York,
  NY, USA, 2016. ACM.

\bibitem{chen2019self}
Yuhua Chen, Cordelia Schmid, and Cristian Sminchisescu.
\newblock Self-supervised learning with geometric constraints in monocular
  video: Connecting flow, depth, and camera.
\newblock In {\em Proceedings of the IEEE International Conference on Computer
  Vision}, pages 7063--7072, 2019.

\bibitem{devlin2018bert}
Jacob Devlin, Ming-Wei Chang, Kenton Lee, and Kristina Toutanova.
\newblock Bert: Pre-training of deep bidirectional transformers for language
  understanding.
\newblock {\em arXiv preprint arXiv:1810.04805}, 2018.

\bibitem{dosovitskiy2020image}
Alexey Dosovitskiy, Lucas Beyer, Alexander Kolesnikov, Dirk Weissenborn,
  Xiaohua Zhai, Thomas Unterthiner, Mostafa Dehghani, Matthias Minderer, Georg
  Heigold, Sylvain Gelly, et~al.
\newblock An image is worth 16x16 words: Transformers for image recognition at
  scale.
\newblock {\em arXiv preprint arXiv:2010.11929}, 2020.

\bibitem{durack2014long}
Paul~J Durack, Susan~E Wijffels, and Peter~J Gleckler.
\newblock Long-term sea-level change revisited: the role of salinity.
\newblock {\em Environmental Research Letters}, 9(11):114017, 2014.

\bibitem{early2022scene}
Joseph Early, Ying-Jung Deweese, Christine Evers, and Sarvapali Ramchurn.
\newblock Scene-to-patch earth observation: Multiple instance learning for land
  cover classification.
\newblock {\em arXiv preprint arXiv:2211.08247}, 2022.

\bibitem{esa_aero}
ESA.
\newblock Aerosols, 2020.
\newblock [Online; accessed: 16.01.2023].

\bibitem{feng2019hypergraph}
Yifan Feng, Haoxuan You, Zizhao Zhang, Rongrong Ji, and Yue Gao.
\newblock Hypergraph neural networks.
\newblock In {\em Proceedings of the AAAI Conference on Artificial
  Intelligence}, volume~33, pages 3558--3565, 2019.

\bibitem{fifty2021efficiently}
Chris Fifty, Ehsan Amid, Zhe Zhao, Tianhe Yu, Rohan Anil, and Chelsea Finn.
\newblock Efficiently identifying task groupings for multi-task learning.
\newblock {\em Advances in Neural Information Processing Systems},
  34:27503--27516, 2021.

\bibitem{gao2022fbsnet}
Guangwei Gao, Guoan Xu, Juncheng Li, Yi Yu, Huimin Lu, and Jian Yang.
\newblock Fbsnet: A fast bilateral symmetrical network for real-time semantic
  segmentation.
\newblock {\em IEEE Transactions on Multimedia}, 2022.

\bibitem{girshick2014rich}
Ross Girshick, Jeff Donahue, Trevor Darrell, and Jitendra Malik.
\newblock Rich feature hierarchies for accurate object detection and semantic
  segmentation.
\newblock In {\em Proceedings of the IEEE conference on computer vision and
  pattern recognition}, pages 580--587, 2014.

\bibitem{gordon2019depth}
Ariel Gordon, Hanhan Li, Rico Jonschkowski, and Anelia Angelova.
\newblock Depth from videos in the wild: Unsupervised monocular depth learning
  from unknown cameras.
\newblock In {\em Proceedings of the IEEE/CVF International Conference on
  Computer Vision}, pages 8977--8986, 2019.

\bibitem{gou2021knowledge}
Jianping Gou, Baosheng Yu, Stephen~J Maybank, and Dacheng Tao.
\newblock Knowledge distillation: A survey.
\newblock {\em International Journal of Computer Vision}, 129(6):1789--1819,
  2021.

\bibitem{goyal2020graph}
Palash Goyal, Sachin Raja, Di Huang, Sujit~Rokka Chhetri, Arquimedes Canedo,
  Ajoy Mondal, Jaya Shree, and CV Jawahar.
\newblock Graph representation ensemble learning.
\newblock In {\em 2020 IEEE/ACM International Conference on Advances in Social
  Networks Analysis and Mining (ASONAM)}, pages 24--31. IEEE, 2020.

\bibitem{guizilini2020semantically}
Vitor Guizilini, Rui Hou, Jie Li, Rares Ambrus, and Adrien Gaidon.
\newblock Semantically-guided representation learning for self-supervised
  monocular depth.
\newblock {\em arXiv preprint arXiv:2002.12319}, 2020.

\bibitem{haller2021self}
Emanuela Haller, Elena Burceanu, and Marius Leordeanu.
\newblock Self-supervised learning in multi-task graphs through iterative
  consensus shift.
\newblock {\em arXiv preprint arXiv:2103.14417}, 2021.

\bibitem{haller2021unsupervised}
Emanuela Haller, Elena Burceanu, and Marius Leordeanu.
\newblock Self-supervised learning in multi-task graphs through iterative
  consensus shift.
\newblock {\em BMVC}, 2021.

\bibitem{he2017unsupervised}
Li He, Xing Xu, Huimin Lu, Yang Yang, Fumin Shen, and Heng~Tao Shen.
\newblock Unsupervised cross-modal retrieval through adversarial learning.
\newblock In {\em 2017 IEEE International Conference on Multimedia and Expo
  (ICME)}, pages 1153--1158. IEEE, 2017.

\bibitem{Hu_2019_CVPR}
Di Hu, Feiping Nie, and Xuelong Li.
\newblock Deep multimodal clustering for unsupervised audiovisual learning.
\newblock In {\em The IEEE Conference on Computer Vision and Pattern
  Recognition (CVPR)}, June 2019.

\bibitem{huh2016makes}
Minyoung Huh, Pulkit Agrawal, and Alexei~A Efros.
\newblock What makes imagenet good for transfer learning?
\newblock {\em arXiv preprint arXiv:1608.08614}, 2016.

\bibitem{jiang2023forkmerge}
Junguang Jiang, Baixu Chen, Junwei Pan, Ximei Wang, Liu Dapeng, Jie Jiang, and
  Mingsheng Long.
\newblock Forkmerge: Overcoming negative transfer in multi-task learning.
\newblock {\em arXiv preprint arXiv:2301.12618}, 2023.

\bibitem{kwiatkowska2002merger}
Ewa~J Kwiatkowska and Giulietta~S Fargion.
\newblock Merger of ocean color information from multiple satellite missions
  under the nasa simbios project office.
\newblock In {\em Proceedings of the Fifth International Conference on
  Information Fusion. FUSION 2002.(IEEE Cat. No. 02EX5997)}, volume~1, pages
  291--298. IEEE, 2002.

\bibitem{leordeanu2021semi}
Marius Leordeanu, Mihai~Cristian P{\^\i}rvu, Dragos Costea, Alina~E Marcu, Emil
  Slusanschi, and Rahul Sukthankar.
\newblock Semi-supervised learning for multi-task scene understanding by neural
  graph consensus.
\newblock In {\em Proceedings of the AAAI Conference on Artificial
  Intelligence}, volume~35, pages 1882--1892, 2021.

\bibitem{li2020nasa}
Alan~S Li, Ved Chirayath, Michal Segal-Rozenhaimer, Juan~L Torres-Perez, and
  Jarrett van~den Bergh.
\newblock Nasa nemo-net's convolutional neural network: Mapping marine habitats
  with spectrally heterogeneous remote sensing imagery.
\newblock {\em IEEE journal of selected topics in applied earth observations
  and remote sensing}, 13:5115--5133, 2020.

\bibitem{Li_2019_CVPR}
Yunzhu Li, Jun-Yan Zhu, Russ Tedrake, and Antonio Torralba.
\newblock Connecting touch and vision via cross-modal prediction.
\newblock In {\em The IEEE Conference on Computer Vision and Pattern
  Recognition (CVPR)}, June 2019.

\bibitem{ling2022semi}
Huaming Ling, Chenglong Bao, Xin Liang, and Zuoqiang Shi.
\newblock Semi-supervised clustering via dynamic graph structure learning.
\newblock {\em arXiv preprint arXiv:2209.02513}, 2022.

\bibitem{liu2019structured}
Yifan Liu, Ke Chen, Chris Liu, Zengchang Qin, Zhenbo Luo, and Jingdong Wang.
\newblock Structured knowledge distillation for semantic segmentation.
\newblock In {\em Proceedings of the IEEE/CVF Conference on Computer Vision and
  Pattern Recognition}, pages 2604--2613, 2019.

\bibitem{marcu2020semantics}
Alina Marcu, Vlad Licaret, Dragos Costea, and Marius Leordeanu.
\newblock Semantics through time: Semi-supervised segmentation of aerial videos
  with iterative label propagation.
\newblock In {\em Asian Conference on Computer Vision (ACCV)}, pages
  2881--2890, 2020.

\bibitem{neo}
Earth~Observations NASA.
\newblock Nasa earth observations dataset, 2020.
\newblock [Online; accessed: 16.01.2023].

\bibitem{ngclib}
NGC.
\newblock Neural graph consensus library, 2022.
\newblock [Online; accessed: 16.01.2023].

\bibitem{pan2004automatic}
Jia-Yu Pan, Hyung-Jeong Yang, Christos Faloutsos, and Pinar Duygulu.
\newblock Automatic multimedia cross-modal correlation discovery.
\newblock In {\em Proceedings of the tenth ACM SIGKDD international conference
  on Knowledge discovery and data mining}, pages 653--658. ACM, 2004.

\bibitem{NEURIPS2019_9015}
Adam Paszke, Sam Gross, Francisco Massa, Adam Lerer, James Bradbury, Gregory
  Chanan, Trevor Killeen, Zeming Lin, Natalia Gimelshein, Luca Antiga, Alban
  Desmaison, Andreas Kopf, Edward Yang, Zachary DeVito, Martin Raison, Alykhan
  Tejani, Sasank Chilamkurthy, Benoit Steiner, Lu Fang, Junjie Bai, and Soumith
  Chintala.
\newblock Pytorch: An imperative style, high-performance deep learning library.
\newblock In {\em Advances in Neural Information Processing Systems 32}, pages
  8024--8035. Curran Associates, Inc., 2019.

\bibitem{yt_consistency_video}
Mihai Pirvu.
\newblock Consistency maps video \url{https://youtu.be/HKWCVzWZ9Rk}.

\bibitem{ranjan2019competitive}
Anurag Ranjan, Varun Jampani, Lukas Balles, Kihwan Kim, Deqing Sun, Jonas
  Wulff, and Michael~J Black.
\newblock Competitive collaboration: Joint unsupervised learning of depth,
  camera motion, optical flow and motion segmentation.
\newblock In {\em Proceedings of the IEEE Conference on Computer Vision and
  Pattern Recognition}, pages 12240--12249, 2019.

\bibitem{rolf2021generalizable}
Esther Rolf, Jonathan Proctor, Tamma Carleton, Ian Bolliger, Vaishaal Shankar,
  Miyabi Ishihara, Benjamin Recht, and Solomon Hsiang.
\newblock A generalizable and accessible approach to machine learning with
  global satellite imagery.
\newblock {\em Nature communications}, 12(1):1--11, 2021.

\bibitem{segal2020cloud}
Michal Segal-Rozenhaimer, Alan Li, Kamalika Das, and Ved Chirayath.
\newblock Cloud detection algorithm for multi-modal satellite imagery using
  convolutional neural-networks (cnn).
\newblock {\em Remote Sensing of Environment}, 237:111446, 2020.

\bibitem{segovia2021does}
Ignacio Segovia~Dominguez, Huikyo Lee, Yuzhou Chen, Michael Garay, Krzysztof~M
  Gorski, and Yulia~R Gel.
\newblock Does air quality really impact covid-19 clinical severity: coupling
  nasa satellite datasets with geometric deep learning.
\newblock In {\em Proceedings of the 27th ACM SIGKDD Conference on Knowledge
  Discovery \& Data Mining}, pages 3540--3548, 2021.

\bibitem{stekovic2020casting}
Sinisa Stekovic, Friedrich Fraundorfer, and Vincent Lepetit.
\newblock Casting geometric constraints in semantic segmentation as
  semi-supervised learning.
\newblock In {\em The IEEE Winter Conference on Applications of Computer
  Vision}, pages 1854--1863, 2020.

\bibitem{stretcu2019graph}
Otilia Stretcu, Krishnamurthy Viswanathan, Dana Movshovitz-Attias, Emmanouil
  Platanios, Sujith Ravi, and Andrew Tomkins.
\newblock Graph agreement models for semi-supervised learning.
\newblock {\em Advances in Neural Information Processing Systems}, 32, 2019.

\bibitem{tosi2020distilled}
Fabio Tosi, Filippo Aleotti, Pierluigi~Zama Ramirez, Matteo Poggi, Samuele
  Salti, Luigi~Di Stefano, and Stefano Mattoccia.
\newblock Distilled semantics for comprehensive scene understanding from
  videos.
\newblock In {\em Proceedings of the IEEE/CVF Conference on Computer Vision and
  Pattern Recognition}, pages 4654--4665, 2020.

\bibitem{tudisco2021nonlinear}
Francesco Tudisco, Konstantin Prokopchik, and Austin~R Benson.
\newblock A nonlinear diffusion method for semi-supervised learning on
  hypergraphs.
\newblock {\em arXiv preprint arXiv:2103.14867}, 2021.

\bibitem{upadhyaya2022multi}
Apoorva Upadhyaya, Marco Fisichella, and Wolfgang Nejdl.
\newblock A multi-task model for sentiment aided stance detection of climate
  change tweets.
\newblock {\em arXiv preprint arXiv:2211.03533}, 2022.

\bibitem{wang2021knowledge}
Lin Wang and Kuk-Jin Yoon.
\newblock Knowledge distillation and student-teacher learning for visual
  intelligence: A review and new outlooks.
\newblock {\em IEEE Transactions on Pattern Analysis and Machine Intelligence},
  2021.

\bibitem{wu2021real}
Jipeng Wu, Rongrong Ji, Jianzhuang Liu, Mingliang Xu, Jiawen Zheng, Ling Shao,
  and Qi Tian.
\newblock Real-time semantic segmentation via sequential knowledge
  distillation.
\newblock {\em Neurocomputing}, 439:134--145, 2021.

\bibitem{xiao2004modeling}
Xiangming Xiao, Qingyuan Zhang, Bobby Braswell, Shawn Urbanski, Stephen Boles,
  Steven Wofsy, Berrien Moore~III, and Dennis Ojima.
\newblock Modeling gross primary production of temperate deciduous broadleaf
  forest using satellite images and climate data.
\newblock {\em Remote sensing of environment}, 91(2):256--270, 2004.

\bibitem{yang2018unsupervised}
Zhenheng Yang, Peng Wang, Wei Xu, Liang Zhao, and Ramakant Nevatia.
\newblock Unsupervised learning of geometry from videos with edge-aware
  depth-normal consistency.
\newblock In {\em Thirty-Second AAAI Conference on Artificial Intelligence},
  2018.

\bibitem{Yeo_2021_ICCV}
Teresa Yeo, O\u{g}uzhan~Fatih Kar, and Amir Zamir.
\newblock Robustness via cross-domain ensembles.
\newblock In {\em Proceedings of the IEEE/CVF International Conference on
  Computer Vision (ICCV)}, pages 12189--12199, October 2021.

\bibitem{zamir2020robust}
Amir~R Zamir, Alexander Sax, Nikhil Cheerla, Rohan Suri, Zhangjie Cao, Jitendra
  Malik, and Leonidas~J Guibas.
\newblock Robust learning through cross-task consistency.
\newblock In {\em Proceedings of the IEEE/CVF Conference on Computer Vision and
  Pattern Recognition}, pages 11197--11206, 2020.

\bibitem{zamir2018taskonomy}
Amir~R Zamir, Alexander Sax, William Shen, Leonidas~J Guibas, Jitendra Malik,
  and Silvio Savarese.
\newblock Taskonomy: Disentangling task transfer learning.
\newblock In {\em Proceedings of the IEEE conference on computer vision and
  pattern recognition}, pages 3712--3722, 2018.

\bibitem{zhang2017split}
Richard Zhang, Phillip Isola, and Alexei~A Efros.
\newblock Split-brain autoencoders: Unsupervised learning by cross-channel
  prediction.
\newblock In {\em Proceedings of the IEEE Conference on Computer Vision and
  Pattern Recognition}, pages 1058--1067, 2017.

\bibitem{zhang2022topformer}
Wenqiang Zhang, Zilong Huang, Guozhong Luo, Tao Chen, Xinggang Wang, Wenyu Liu,
  Gang Yu, and Chunhua. Shen.
\newblock Topformer: Token pyramid transformer for mobile semantic
  segmentation.
\newblock {\em Proc. IEEE Conf. Computer Vision and Pattern Recognition
  (CVPR)}, 2022.

\bibitem{zhao2018sound}
Hang Zhao, Chuang Gan, Andrew Rouditchenko, Carl Vondrick, Josh McDermott, and
  Antonio Torralba.
\newblock The sound of pixels.
\newblock In {\em Proceedings of the European Conference on Computer Vision
  (ECCV)}, pages 570--586, 2018.

\bibitem{zhen2020joint}
Mingmin Zhen, Jinglu Wang, Lei Zhou, Shiwei Li, Tianwei Shen, Jiaxiang Shang,
  Tian Fang, and Long Quan.
\newblock Joint semantic segmentation and boundary detection using iterative
  pyramid contexts.
\newblock In {\em Proceedings of the IEEE/CVF Conference on Computer Vision and
  Pattern Recognition}, pages 13666--13675, 2020.

\bibitem{zhou2017unsupervised}
Tinghui Zhou, Matthew Brown, Noah Snavely, and David~G Lowe.
\newblock Unsupervised learning of depth and ego-motion from video.
\newblock In {\em Proceedings of the IEEE Conference on Computer Vision and
  Pattern Recognition}, pages 1851--1858, 2017.

\end{thebibliography}
}

\end{document}